\documentclass[conference]{IEEEtran}
\IEEEoverridecommandlockouts
\usepackage{cite}
\usepackage{amsmath,amssymb,amsfonts}
\usepackage{algorithmic}
\usepackage{graphicx}
\usepackage{dblfloatfix}
\usepackage{subcaption}
\usepackage{textcomp}
\usepackage[table]{xcolor}
\usepackage{xcolor}
\def\BibTeX{{\rm B\kern-.05em{\sc i\kern-.025em b}\kern-.08em
    T\kern-.1667em\lower.7ex\hbox{E}\kern-.125emX}}

\usepackage{tabularx}
\usepackage{url}
\usepackage{caption}

\usepackage{dblfloatfix}

\usepackage{soul}
\usepackage{courier}
\usepackage[T1]{fontenc}
\usepackage{epstopdf}
\newcolumntype{Y}{>{\centering\arraybackslash}X}
\usepackage{collcell}
\usepackage{xfp}

\usepackage{listings}                       

\makeatletter
\newcommand\fs@spaceruled{\def\@fs@cfont{\bfseries}\let\@fs@capt\floatc@ruled%
  \def\@fs@pre{\vspace{0.5\baselineskip}\hrule height.8pt depth0pt \kern2pt}%
  \def\@fs@post{\kern2pt\hrule\relax}%
  \def\@fs@mid{\kern2pt\hrule\kern2pt}%
  \let\@fs@iftopcapt\iftrue}%

\newcommand{\linebreakand}{%
  \end{@IEEEauthorhalign}
  \hfill\mbox{}\par
  \mbox{}\hfill\begin{@IEEEauthorhalign}
}
\makeatother

\begin{document}

\title{Augmented Reality Appendages for Robots: \\ Design Considerations and Recommendations for\\ Maximizing Social and Functional Perception
\thanks{*Equal Contribution}
}

\author{
  \IEEEauthorblockN{Ipek Goktan*}
  \IEEEauthorblockA{\textit{Viterbi School of Engineering} \\
  \textit{Dept. of Computer Science} \\
    \textit{University of Southern California}\\
    Los Angeles, USA \\
    ipekg@usc.edu}
  \and
  \IEEEauthorblockN{Karen Ly*}
  \IEEEauthorblockA{\textit{Viterbi School of Engineering} \\
  \textit{Dept. of Computer Science} \\
    \textit{University of Southern California}\\
    Los Angeles, USA \\
    karenly@usc.edu}
  \and
    \IEEEauthorblockN{Thomas R. Groechel*}
  \IEEEauthorblockA{\textit{Viterbi School of Engineering} \\
  \textit{Dept. of Computer Science} \\
    \textit{University of Southern California}\\
    Los Angeles, USA \\
    groechel@usc.edu}
\linebreakand 
 \IEEEauthorblockN{Maja J. Matari\'c}
 \textit{Dept. of Computer Science} \\
  \IEEEauthorblockA{\textit{Viterbi School of Engineering} \\
    \textit{University of Southern California}\\
    Los Angeles, USA \\
    mataric@usc.edu}
}


\newcommand\spacesize{0.9}
\maketitle

\begin{abstract}
In order to address the limitations of gestural capabilities in physical robots, researchers in Virtual, Augmented, Mixed Reality Human-Robot Interaction (VAM-HRI) have been using augmented-reality visualizations that increase robot expressivity and improve user perception (e.g., social presence). While a multitude of virtual robot deictic gestures (e.g., pointing to an object) have been implemented to improve interactions within VAM-HRI, such systems are often reported to have tradeoffs between functional and social user perceptions of robots, creating a need for a unified approach that considers both attributes. We performed a literature analysis that selected factors that were noted to significantly influence either user perception or task efficiency and propose a set of design considerations and recommendations that address those factors by combining anthropomorphic and non-anthropomorphic virtual gestures based on the motivation of the interaction, visibility of the target and robot, salience of the target, and distance between the target and robot. The proposed recommendations provide the VAM-HRI community with starting points for selecting appropriate gesture types for a multitude of interaction contexts.

\end{abstract}

\begin{IEEEkeywords}
Augmented, Mixed, and Virtual Reality for Human-Robot Interaction;  social perception; functional perception; deictic gesture
\end{IEEEkeywords}


\section{Introduction}
\label{sec:introduction}

To address the limitations of gestural capabilities in physical robots, researchers in the field of Virtual, Augmented, Mixed Reality Human-Robot Interaction (VAM-HRI) have been using augmented reality visualizations to enhance robot expressivity and improve user perception of robots (e.g., social presence, functional capability) \cite{cha2018survey}. Research has shown that gestures can be designed to boost users' functional or social perception of robots. \textit{Functional perception} is the user’s belief that a robot is able to accurately communicate its functional capabilities and intentions (e.g., the user can easily choose the target from a list of objects) \cite{groechel2019using}, \cite{hamilton2020tradeoffs}. \textit{Social perception} is the user's belief that a robot is capable of participating in the interaction as a social actor that is an interactive, autonomous, and adaptable agent that performs anthropomorphic actions \cite{jackson2021theory}.

\begin{figure}[t]
    \begin{subfigure}{\columnwidth}
      \centering
      \includegraphics[width=1\linewidth]{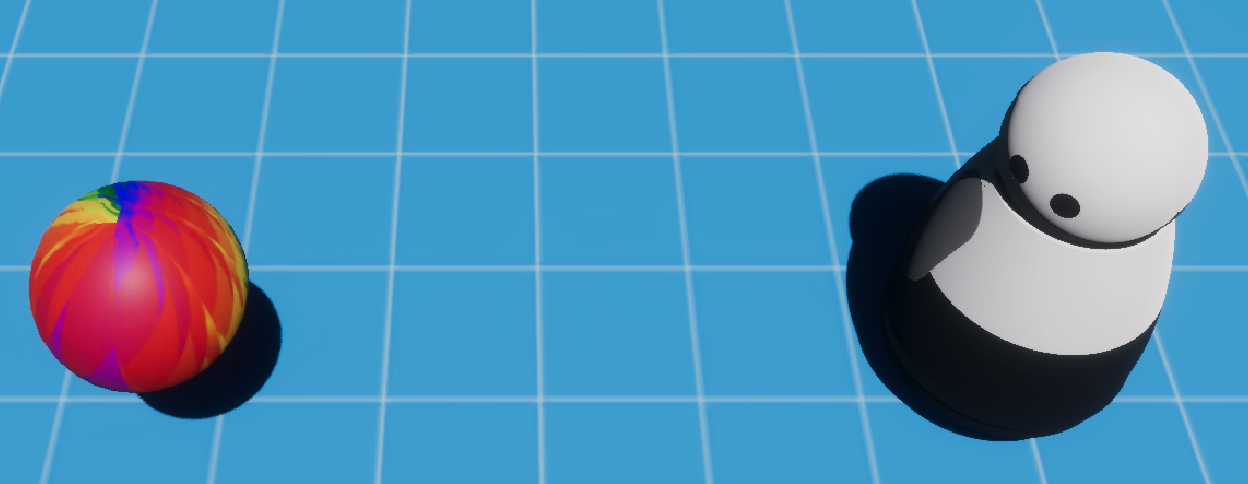}
      \label{fig:sub:firstfirst}
    \end{subfigure}%
    
    \begin{subfigure}{\columnwidth}
      \centering
      \includegraphics[width=1\linewidth]{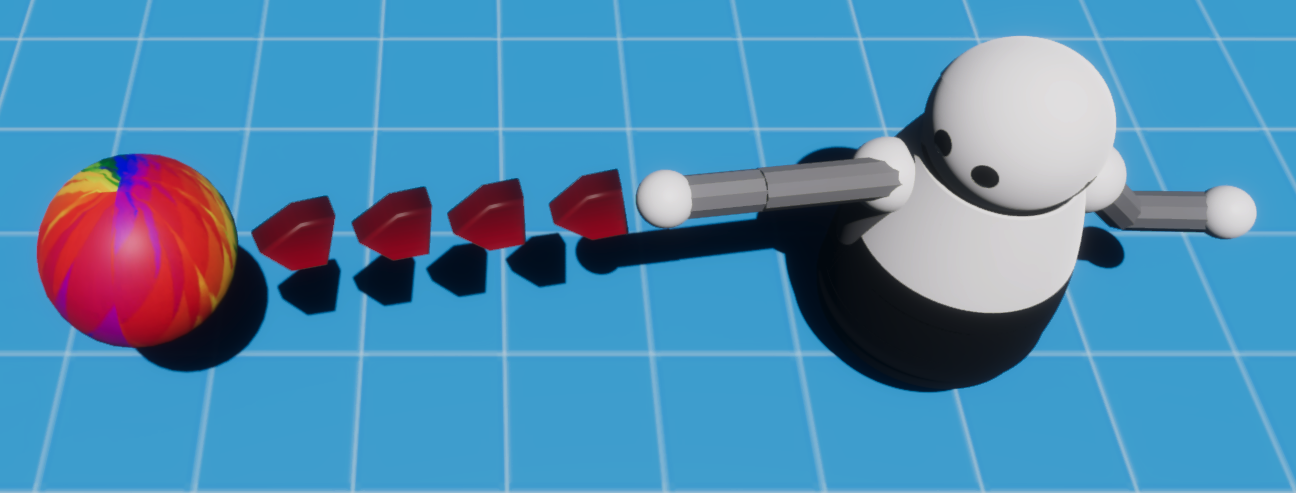}
      \label{fig:sub:firstsecond}
    \end{subfigure}%
    \caption{Kuri indicating a target object for the user to attend to. \textbf{Top}: Kuri without augmented reality (AR) additions. \textbf{Bottom}: Kuri gestures at a target object using combined anthropomorphic augmented reality appendages (arms) and non-anthropomorphic gestures (arrows).}
    \label{fig:sub:bothanthroandnonanthro}
\end{figure}%

\begin{figure*}[t!]
    \begin{subfigure}{0.27\textwidth}
      \centering
      \includegraphics[width=\spacesize\textwidth]{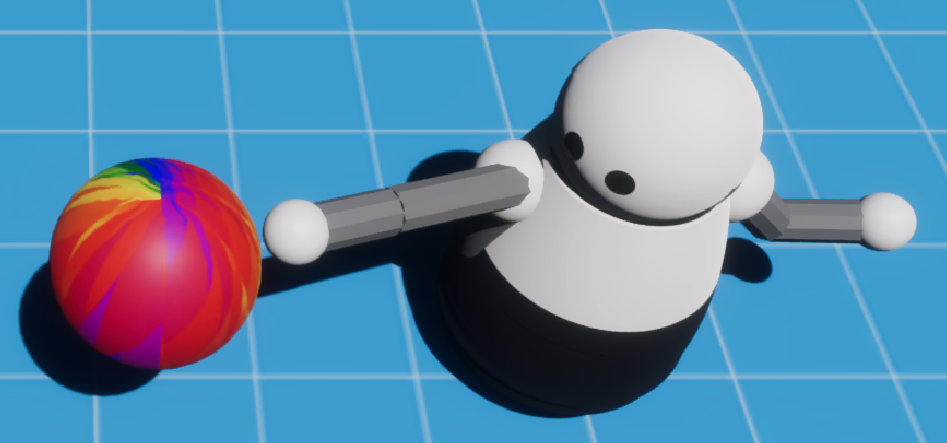}
      \caption{Anthropomorphic}
      \label{fig:sub:anthro}
    \end{subfigure}%
    \begin{subfigure}{0.36\textwidth}
      \centering
      \includegraphics[width=\spacesize\textwidth]{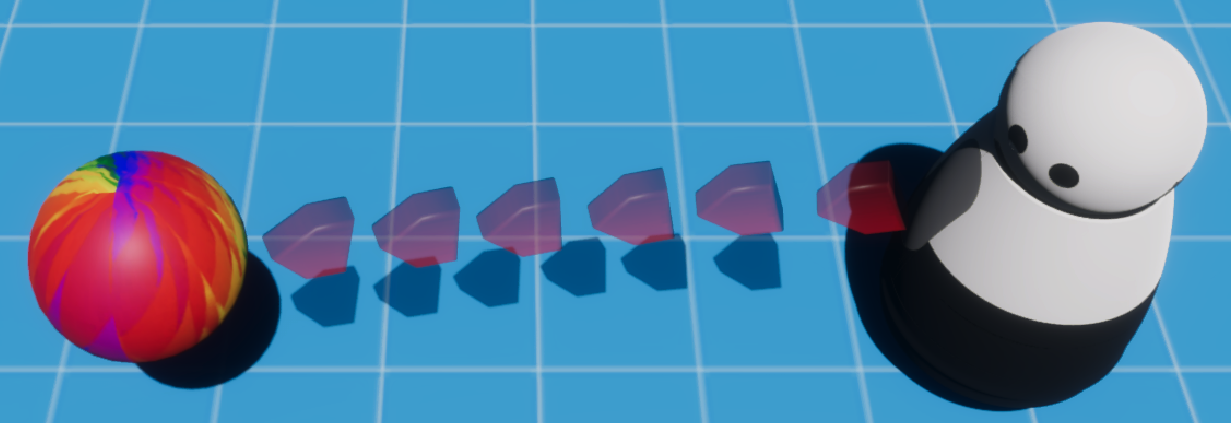}
      \caption{Non-Anthropomorphic}
      \label{fig:sub:nonanthro}
    \end{subfigure}
    \begin{subfigure}{0.37\textwidth}
      \centering
      \includegraphics[width=\spacesize\textwidth]{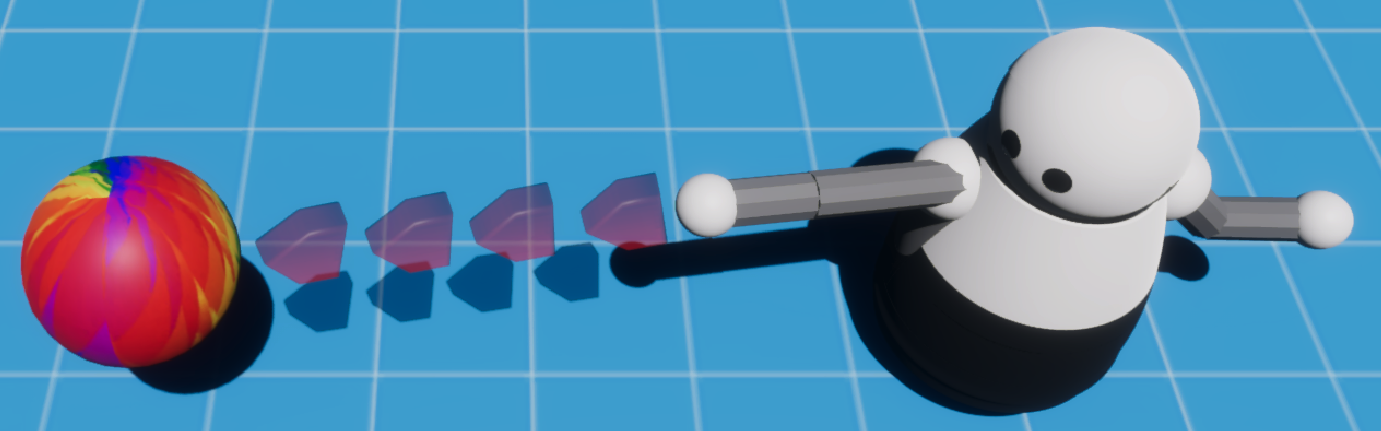}
      \caption{Anthropomorphic and Non-Anthropomorphic}
      \label{fig:sub:combined}
    \end{subfigure}%
    \caption{Kuri robot gestures toward the target sphere by (a) pointing with projected augmented reality arms, (b) using an arrow line, and (c) pointing with projected augmented reality arms and an arrow line. Compared to Fig. \ref{fig:sub:bothanthroandnonanthro}, the arrow line is transparent thus obfuscating less of the background.}
\end{figure*}

A virtual gesture performed by a robot can be either anthropomorphic or non-anthropomorphic. Anthropomorphic gestures, such as those in Fig. \ref{fig:sub:anthro}, are human-like gestures \cite{erel2018interpreting}, \cite{wadgaonkar2021exploring} that support Human-Robot Interaction (HRI) \cite{hamilton2020tradeoffs}. Non-anthropomorphic gestures are abstract gestures (e.g., virtual arrows) that aim to communicate goals of actions efficiently and accurately \cite{erel2018interpreting}, \cite{hamilton2020tradeoffs}. While non-anthropomorphic gestures can be 2D (e.g., a flat dashed line) or 3D (e.g., arrow line in Fig. \ref{fig:sub:nonanthro}), our proposed recommendations suggest using 3D non-anthropomorphic gestures because they situate gestures in real interaction space and communicate added information such as directionality and depth (e.g., via shading/shadows). Further, utilizing 3D non-anthropomorphic gestures may increase social perceptions of the robot as users may associate the non-anthropormorphic signal as part of the robot. Additionally, anthropomorphic and non-anthropomorphic gestures can be used in tandem, as shown in Fig. \ref{fig:sub:bothanthroandnonanthro}, where the robot points at an object and an arrow connects its virtual arm to the object.

While a variety of virtual robot deictic gestures (i.e., gestures that point toward objects or areas in order to redirect user's attention) have been implemented to improve VAM-HRI, both anthropomorphic and non-anthropomorphic gesture systems have been found to involve tradeoffs between functional task performance and user social perception of robots \cite{hamilton2020tradeoffs}, creating a need for a unified approach that considers both functional and social attributes. 

Toward that goal, we propose a set of design recommendations and techniques that combine anthropomorphic and non-anthropomorphic virtual gestures based on the placement of the robot, user, and target object during an interaction. Our design recommendations were compiled by analyzing the relevant literature for design considerations, and highlighting factors that were reported to significantly influence either social perception or task efficiency \cite{hamilton2020tradeoffs}, \cite{walker2018communicating}, \cite{stogsdill2021pointless}, \cite{piwek2009salience}, \cite{tran2021robot}. We distil the  findings into the following key factors: motivation of the interaction, a target's visibility, a target's salience, and a target's distance. To illustrate the recommendations, we consider the example of a robot attempting to draw attention to an object via an anthropomorphic deictic gesture and use the proposed recommendations to create AR gestures that consider both functional and social user perceptions.

\section{Background}
\label{sec:Background}
This paper builds on past work on communicating robot intent, robot gestures, anthropomorphism, and expressivity in VAM-HRI, briefly summarized next. 

\subsection{Influencing Functional Perception}
Functional perception is a key concept in Augmented Reality (AR). Developers often project AR on objects in the physical world in order to quickly and concisely communicate concepts.  Walker et al. \cite{walker2018communicating} designed and evaluated various explicit and implicit AR interfaces and reported that certain designs (e.g., NavPoints) were able to significantly improve the communication of robot intent as well as objective task efficiency. Other studies have also shown that utilizing AR can increase user task speed and facilitate human-robot collaboration \cite{tran2021robot}, \cite{bagchi2018towards}, \cite{chandan2021arroch}, \cite{krenn2021s}. These findings suggest that AR visualizations can be used to accurately communicate robot intent within a desirable reaction time, and therefore boost the functional perception of a robot. 

\subsection{Influencing Social Perception}
AR has also been used as a medium to enhance the expressivity of mobility-constrained robots (i.e., robots that are either unable to move or move slowly in the context of the interaction) and give them a prominent role as social actors within interactions \cite{groechel2019using} \cite{young2007cartooning}. Additionally, past research suggests that implementing AR arms on a low-expressivity robot can increase its physical and social presence. Groechel et al. \cite{groechel2019using} projected AR arms on a physical robot and found that participants were more likely to view the robot as a physical teammate. VAM-HRI has explored other techniques for using augmented reality to boost expressivity, such as ``robot expressionism through cartooning,'' which applies common comic and cartoon art  \cite{young2007cartooning}. Social perception is also highly influenced by the user's action attribution of the manipulation (e.g., when an object moves during an interaction, the user attributes the movement to the robot) \cite{groechel2022tool}. The intended anchor location perception being robot-anchored increases social perception.

\subsection{Trade-Offs Between Functional and Social Perception}
Although Walker et al. \cite{walker2018communicating} reported an improvement in the communication of robot intent and task efficiency, they also observed trade-offs between intent clarity and users' perception of the robot as a teammate. This trade-off between social and functional perception has been recognized in other work as well. Hamilton et al. \cite{hamilton2020tradeoffs} compared  ego-sensitive-gestures (e.g., a virtual arrow placed on a robot) and non-ego-sensitive allocentric gestures (e.g., a virtual arrow placed on a target), and found that using non-ego-sensitive gestures resulted in faster reaction time and greater accuracy, whereas ego-sensitive gestures resulted in higher social perception and likability \cite{hamilton2020tradeoffs}. 

\section{Design Considerations and Recommendations}
\label{sec:Design Considerations and Recommendations}

Research has shown that when discussing an object, people consider the context of the target object and the surrounding environment when making the decision between using deictic and non-deictic gestures to draw attention to the it \cite{stogsdill2021pointless}.
Similar to how people take account of contextual factors when determining when and how to use gestures, we highlight four main design considerations for the VAM-HRI community, based on a literature review \cite{hamilton2020tradeoffs}, \cite{walker2018communicating}, \cite{stogsdill2021pointless}, \cite{piwek2009salience}, \cite{tran2021robot}. From these design considerations, a set of proposed recommendations were developed. The recommendations were designed for mobility-constrained robots with an AR field of view. A summary of the recommendations is found in Table \ref{tab:tech-approach-table}, which shows the proposed method of gesturing for all combinations of design considerations. 

\newcommand\colsize{0.11925} 
\begin{table*}[b!] 
\centering
\caption{Recommended Gesture Type Based on Design Considerations: \textbf{"A"} is  anthropomorphic gesture and \textbf{"NA"} is non-anthropomorphic gesture. \textbf{"NA*"} suggests a directional non-anthropomorphic gesture (e.g., vector with arrow pointing in the direction of the target). The ordering between \textbf{"A"} and \textbf{"NA}" suggests the prioritization between the two types. }
\label{tab:tech-approach-table}
\rowcolors{2}{white}{gray!25}
\begin{tabularx}{\textwidth}{|>{\hsize=\colsize\textwidth}X|>{\hsize=\colsize\textwidth}Y>{\hsize=\colsize\textwidth}Y>{\hsize=\colsize\textwidth}Y>{\hsize=\colsize\textwidth}Y>{\hsize=\colsize\textwidth}Y>{\hsize=\colsize\textwidth}Y|}
\hline
\textbf{} & \textbf{Functional Motivation, High Salience} &  \textbf{Functional Motivation, Low Salience} &  \textbf{Social Motivation, High Salience} &  
\textbf{Social Motivation, Low Salience} &  
\textbf{Functional and Social Motivation, High Salience} &  \textbf{Functional and Social Motivation, Low Salience} \\
\hline
\textbf{Close to Target, Both in FOV}    & A      & NA + A & A       & A + NA  & A      & A + NA \\
\textbf{Close to Target, Only one in FOV} & NA     & NA     & A + NA  & NA + A  & NA + A & NA + A \\
\textbf{Close to Target, Neither in FOV}  & NA*    & NA*    & A + NA* & A + NA* & NA*    & NA*    \\
\textbf{Far from Target. Both in FOV}    & NA + A & NA + A & A + NA  & NA + A  & A + NA & NA + A \\
\textbf{Far from Target Only one in FOV} & NA     & NA     & NA + A  & NA + A  & NA + A & NA     \\
\textbf{Far from Target, Neither in FOV}  & NA*    & NA*    & NA* + A & NA* + A & NA*    & NA*  \\
\hline
\end{tabularx}
\end{table*}

\subsection{Motivation of the Interaction}
The {\it motivation of the interaction} is the intended end goal or purpose of the interaction. We refer to an {\it interaction} as a single correspondence between a robot and human, and a {\it composite interaction} as sequence of multiple interactions. The motivation of the interaction is for the robot to communicate information to the user in order to achieve a given goal. More specifically, motivation can have a functional component, social component, or both. 
\subsubsection{Motivation as a Functional Component}
An interaction is motivated by a functional component when the objective is the overall task performance. The accuracy and efficiency of how information is conveyed is prioritized, placing the user’s functional perception of the robot above the social perception.

\subsubsection{Motivation as a Social Component}
An interaction contains a social component when the intent of doing the action is for the robot to be perceived as more social. In other words, the user believes the robot has social agency within an interaction \cite{jackson2021theory}, potentially viewing it as acting more human-like than machine-like. Increasing social perception is an example of social motivation. Social perception is the user's belief that a robot is a social agent; the belief may not accurately reflect the robot's true role or capabilities within an interaction.

\textbf{Proposed Recommendations}
\subsubsection*{Functional Component} 
If the motivation for a human-robot interaction contains only a functional component (e.g., increasing functional perception), prior research suggests utilizing a non-anthropomorphic gesture to maximize the efficiency and accuracy of task performance \cite{hamilton2020tradeoffs}, \cite{tran2021robot}. Similarly, Krenn et al.  \cite{krenn2021s} found that robot deictic gestures can help the user interpret the next intended task. Given that the social presence of the robot is not prioritized, solely using non-anthropomorphic gestures will ensure clear and efficient communication \cite{hamilton2020tradeoffs}. 
\subsubsection*{Social Component}
If the motivation of a human-robot interaction contains only a social component (e.g., increasing social perception), prior research suggests including anthropomorphic gestures (e.g., virtual arm for pointing). Deictic gesturing, for example, portrays the robot as a social agent that is an active part of the interaction \cite{hamilton2020tradeoffs}. Using this type of gesture solidifies the robot as an agent and encourages the user to refer back to the robot, increasing the robot's salience within the interaction \cite{hamilton2020tradeoffs}. While anthropomorphic gestures provide robots with the communication and social presence of human-like gesturing, they impose limitations in terms of communication speed and accuracy. Physically-realistic gesturing takes more time than displaying an arrow on the screen. This phenomenon increases social perception but hinders functional perception due to the reduction in the speed of communication.
\subsubsection*{Functional and Social Component} 
If the motivation consists both of functional and social components, then non-anthropomorphic and anthropomorphic gestures can be used in combination, as illustrated in Fig. \ref{fig:sub:combined}, but the combination is not always the best approach. For consistency, it may be best to use anthropomorphic gestures and consider including non-anthropomorphic gestures only when needed for speed or clarity. The goal of a composite interaction may not always be to maximize functional and social components. For example, if the interaction requires high functional capability, it may be advantageous to forgo anthropomorphic gesturing altogether.

\subsection{Visibility}
Visibility refers to whether the user can see the target object in their field of view (FOV) \cite{stogsdill2021pointless}. This category can be affected by virtual or physical occlusions. During an interaction, the user may see both the robot and target (AND), only the robot or target (XOR), or neither (NAND). 

\begin{figure}[h]
      \centering
      \includegraphics[width=\columnwidth]{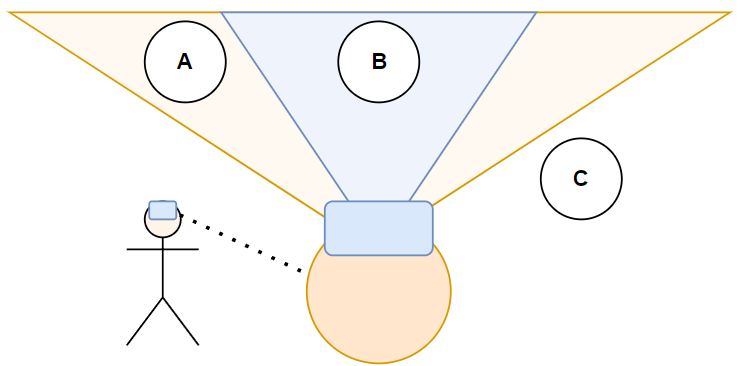}
      \caption{Top down view of a user wearing an AR head-mounted display. Object \textbf{A} is in the user's field of view (FOV) but not the AR FOV. Object \textbf{B} is in both views. Object \textbf{C} is in neither. }
      \label{fig:fov}
\end{figure}%

This section assumes that an object within the FOV is within the boundaries of the AR FOV (e.g., object B in Fig. \ref{fig:fov}). AR FOV is a subset of human FOV that is limited by the scale in which the platform displaying AR components is able to capture the environment the user sees. A physical object  that is in the user's FOV but outside of the AR FOV is considered to be out of frame. This includes the robot as long as the AR appendages are not visible within the AR FOV. Because the FOV of AR devices is dynamically changing and interaction-dependent, our proposed recommendations specifically target the use of an AR FOV rather than a human FOV. Consequently, the recommendations for this category may not be applicable if a non-AR FOV is being used.

\subsubsection{Both Robot and Target Object are Visible (AND)}
The user can clearly view the robot and the target object in their FOV without having to adjust their gaze.
\subsubsection{Visible Robot or Object but Not Both (XOR)}
The user can only view the robot or only view the target object in their FOV. The user may need to adjust their gaze in order to see both targets at once or it may not be possible for them to be in the same FOV.
\subsubsection{Neither Robot nor Target Object is Visible (NAND)}
The user cannot see the robot or the target object because they are placed outside of the viewer’s FOV. 
\newpage
\textbf{Proposed Recommendation}
\subsubsection*{Both Target and Robot in User’s FOV (AND)}
Humans are more likely to use deictic gestures (e.g., pointing) when the target is visible \cite{stogsdill2021pointless}. To maximize the robot's social perception, if both the target object and the robot are within the user’s FOV, prior research suggests to use anthropomorphic gestures, as shown in Fig. \ref{fig:sub:anthro}. The robot using anthropomorphic deictic gestures will improve the user's recall and human-robot rapport \cite{stogsdill2021pointless}.

\subsubsection*{Visible Robot or Object but Not Both (XOR)} 
If the target is outside of the user’s FOV but the robot is within it, an anthropomorphic and non-anthropomorphic gesture (e.g., an arrow from the robot’s pointing finger) should be considered to associate the robot to the target. This implementation increases both functional and social perception because anthropomorphic deictic gesturing draws attention and connects the robot’s intentions with the specific task and functionality \cite{stogsdill2021pointless}. The non-anthropomorphic gesture contributes to the functional perception by allowing the user to accurately distinguish the location of the object. The same recommendation applies if user can only see the target. 

In a scenario where only the target is visible, developers can choose whether or not to implement non-anthropomorphic gestures. If non-anthropomorphic gestures are present, the user can clearly identify the target without having to refer back to the robot, increasing functional perception. However, if only anthropomorphic gestures are used, the user will be compelled to find the robot. Looking back at the robot in order to identify its anthropomorphic gestures will make the robot a more prominent as an active agent in the interaction, promoting its social perception.

\subsubsection*{Neither in User’s FOV (NAND)}
If neither the target nor robot are in the user’s FOV, we propose to first consider what the motivation for the interaction is and whether it consists of a functional and/or social component. For instance, one might use a directional
non-anthropomorphic gesture, such as a large arrow that points to the direction of the robot’s location or an arrow as in Fig. \ref{fig:sub:bothanthroandnonanthro}  to redirect the user’s FOV to the robot. However, if the motivation is social, using anthropomorphic deictic gesturing that encourages the user to find the robot first and then search for the object may be more effective for social perception.

\subsection{Salience of Target}
Salience refers to how noticeable a target is within its environment, a property analogous to accessibility \cite{piwek2009salience}. Gestures made toward an object can change its salience value. 

\begin{figure}[h]
      \centering
      \includegraphics[width=\columnwidth]{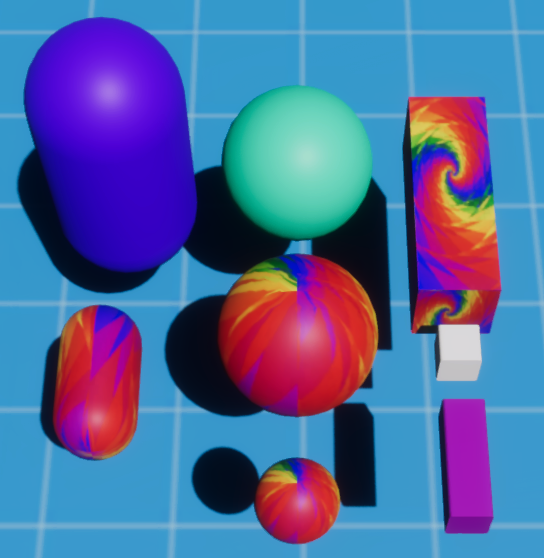}
      \caption{The target object's salience depends on the characteristics of the objects around it (e.g., color, scale, shape). The more the objects share similar characteristics, the lower the target object's salience.}
      \label{fig:saliency}
\end{figure}%

When there are multiple objects in close proximity (e.g., Fig. \ref{fig:saliency}), each may have attributes that distinguish it from others, (e.g., color, scale, shape) and increases its salience \cite{piwek2009salience}. Non-verbal robots are often unable to or highly constrained in their ability to indicate the location of a specific object while utilizing existing salience factors \cite{cha2018survey}; they are typically limited to gesturing toward the object to draw attention to it. Therefore, when it comes to nonverbal signaling, gesturing can increase the salience of an object, but anthropomorphic deictic gesturing often cannot distinguish the object if it is in close proximity to others.

Anthropomorphic deictic gestures can also increase the salience of surrounding objects by bringing attention to the overall area. This is known as implied spatial salience \cite{piwek2009salience}. It is necessary to consider salience when choosing to include anthropomorphic and non-anthropomorphic gestures because it is also determined by how distinguishable the target is from its neighbors.

\textbf{Proposed Recommendation}
\subsubsection*{Target is Far Away from Other Objects}
When the target object in an interaction is not in close proximity to other objects, the user can clearly interpret the intent of an anthropomorphic gesture made toward that object \cite{piwek2009salience}. If there are extraneous factors that prevent the user from clearly identifying the object (e.g., distance, FOV), non-anthropomorphic gestures can be implemented to more accurately communicate the location of the target.

\subsubsection*{Target is in Close Proximity with Other Objects}
The target has low salience if it is in close proximity with other objects \cite{piwek2009salience}. The user may find it difficult to distinguish between the objects in the direction the robot is pointing, and may require additional visual assistance. Therefore, non-anthropomorphic gestures, along with anthropomorphic gestures, can be used to clearly indicate the target. If the motivation of the interaction is functional, developers may avoid anthropomorphic gestures and simply use non-anthropomorphic gestures to quickly indicate the target.

\textbf{Salience of Augmented Reality Components: }
While our design recommendations emphasize the salience of the target object, the salience of the non-anthropomorphic gestures should also be considered. Factors such as color, size, and opacity may influence how the user perceives the robot's gesture. Salience of a non-anthropomorphic gesture affects how disruptive it is to the user, which determines how noticeable it is \cite{parsch2019designing} and how quickly the user can take action after the gesture is produced. For instance, developers generating an arrow to gesture towards a target object may choose to select an opaque color, as in Fig. \ref{fig:sub:bothanthroandnonanthro}, to make it more noticeable, or select a shade that is more transparent, as in Fig.  \ref{fig:sub:combined}, to make it more cohesive with the robot and obstruct less of the background. Considering the opacity, scale, and color of non-anthropomorphic gestures is important because overlaying AR visualizations onto the real world may distract the user and obstruct their FOV \cite{parsch2019designing}. Therefore, non-anthropomorphic gestures that have higher opacity, large scale, or similar color to the background are undesirable. 

Choosing the color of the non-anthropomorphic signal objects (e.g., arrows) can be specifically targeted for different use cases. A signal can be given the same color as the target object for better salience with that object. To avoid blending in with the background, a dynamic coloring system can be created by contrasting the color of the signal's background. A color can be generated as a preprogrammed map (e.g., complimentary colors),  inverted, or changed to a secondary color choice. Finally, when aiming for the most social signal, developers should choose a color on the robot or average color of the robot to boost the signal attribution toward the robot. The same recommendation applies to choosing colors for adding appendages to the robot.

\subsection{Distance}
In an interaction composed of a user, target object, and robot, our distance model prioritizes the displacement between the robot and the object. Hamilton et al. \cite{hamilton2020tradeoffs} reported that there was no significant evidence that the distance between a user and target influenced the social presence of a robot. The findings of their study suggest that there may be a correlation between the robot-target distance and the robot's social presence, which calls for further study. This paper focuses on mobility-constrained robots; distance issues are more complex for highly mobile robots (e.g., drones) that can move quickly and efficently toward a target.

\textbf{Proposed Recommendation: }
\subsubsection*{Target Close to Robot}
If the target object is in close proximity to the robot, anthropomorphic gestures may be used to promote the robot’s social presence. Using diectic gestures would enhance the robot's anthropomorphism because humans tend to use the same gesturing model when objects are visible and in close proximity \cite{stogsdill2021pointless}. The close proximity will also reduce issues of salience \cite{hamilton2020tradeoffs}. Solely using anthropomorphic gestures when the robot is close to the target boosts the social perception of the robot. However, if there are extraneous factors (e.g., salience, field of view) that prevent the user from clearly distinguishing the target object, non-anthropomorphic gestures may be helpful.

\subsubsection*{Target Far from Robot}
If the target object is outside of a predetermined distance boundary (calculated for each interaction considering the specific robot and environment), it can be considered far from the robot. Since distance between the robot and target make it difficult to interpret the direction of deictic gestures, developers may use non-anthropomorphic gestures \cite{stogsdill2021pointless} along with anthropomorphic gestures to indicate the location of a target object without sacrificing functional or social perception. If the interaction is motivated by a functional component, developers may use only non-anthropomorphic gestures in order to promote speed and accuracy \cite{hamilton2020tradeoffs}.

\section{Discussion and Limitations}
\label{Discussion and Limitations}
In this paper, we presented design recommendations that maximize social and functional perception in nonverbal robot gesturing (Table \ref{tab:tech-approach-table}). From prior works, we gathered motivation, visibility, salience, and distance as the key design considerations. For the combination of each consideration, we developed recommendations for whether the robot should indicate the location of an object using anthropomorphic, non-anthropomorphic, or both anthropomorphic and non-anthropomorphic gestures. 

The presented recommendations require further investigation and empirical evaluation. As noted above, they may not be applicable to robots with high mobility (e.g., drones). 

While this paper attempted to introduce a design framework for selecting gesture types in VAM-HRI, gestures are highly dependent on the context of a given interaction (e.g., motivation of an interaction that is not clearly distinguishable as social or functional). Therefore, the design recommendations merit empirical testing and analysis in different scenarios. Future work should explore how various anthropomorphic and non-anthropomorphic gestures (e.g., circling the target object versus generating an arrow line connecting the robot to the target) influence user perception.

\section*{Acknowledgment}
This work was supported by the National Science Foundation (NSF) under award IIS-1925083 and University of Southern California Viterbi School of Engineering under the Merit Research Award and Viterbi Fellowship Award. 

\addtolength{\textheight}{-15cm}

\bibliographystyle{ieeetr}
\bibliography{bibliograph}
\end{document}